\documentclass[conference]{IEEEtran}
\IEEEoverridecommandlockouts
\usepackage{cite}
\usepackage{amsmath,amssymb,amsfonts}
\usepackage{algorithm}
\usepackage{algorithmic}
\usepackage{graphicx}
\usepackage{textcomp}
\usepackage{xcolor}

\usepackage{comment}
\usepackage{subfloat}
\usepackage{subcaption}
\usepackage{booktabs}

\newcommand{\hx}[1]{\textcolor{blue}{[HX: #1]}}

\begin{document}

\title{SwiftBot: A Decentralized Platform for LLM-Powered Federated Robotic Task Execution}

\author{
\IEEEauthorblockN{
YueMing Zhang\IEEEauthorrefmark{1},
Shuai Xu\IEEEauthorrefmark{2},
Zhengxiong Li\IEEEauthorrefmark{3},
Fangtian Zhong\IEEEauthorrefmark{4},
Xiaokun Yang\IEEEauthorrefmark{5},
Hailu Xu\IEEEauthorrefmark{1}
}

\IEEEauthorblockA{
\IEEEauthorrefmark{1}California State University, Long Beach, USA\quad
\IEEEauthorrefmark{2}Case Western Reserve University, USA \quad \\
\IEEEauthorrefmark{3}University of Colorado Denver, USA\quad
\IEEEauthorrefmark{4}Montana State University, USA\quad
\IEEEauthorrefmark{5}University of Houston Clear Lake, USA \\
Emails:\{simon.zhang01@student.csulb.edu, shuai.xu2@case.edu, 
zhengxiong.li@ucdenver.edu, \\ yangxia@uhcl.edu, 
fangtian.zhong@montana.edu, hailu.xu@csulb.edu\}
}
}

\maketitle
\begin{abstract}

%Multi-robot task execution systems struggle to bridge natural language instructions to robot control while efficiently managing distributed resources. Existing approaches face three limitations: rigid hand-coded planners requiring extensive domain engineering, centralized coordination architectures that limit scalability as robot sizes grow, and naive resource management failing to adapt to dynamic workload patterns in distributed edge environments. We present SwiftBot, a decentralized platform integrating LLM-based task decomposition with intelligent container orchestration over a DHT overlay. SwiftBot achieves 94.3\% decomposition accuracy across diverse tasks, reduces task startup latency by 1.5-5.4× and average training latency by 1.4-2.5×, and achieves 1.2-4.7× tail latency under high task load through distributed warm container migration. Evaluation on multimedia tasks validates that co-designing semantic understanding and resource management enables both flexibility and efficiency for real-time robotic task control.

Federated robotic task execution systems require bridging natural language instructions to distributed robot control while efficiently managing computational resources across heterogeneous edge devices without centralized coordination. Existing approaches face three limitations: rigid hand-coded planners requiring extensive domain engineering, centralized coordination that contradicts federated collaboration as robots scale, and static resource allocation failing to share containers across robots when workloads shift dynamically. We present SwiftBot, a federated task execution platform that integrates LLM-based task decomposition with intelligent container orchestration over a DHT overlay, enabling robots to collaboratively execute tasks without centralized control. SwiftBot achieves 94.3\% decomposition accuracy across diverse tasks, reduces task startup latency by 1.5-5.4$\times$ and average training latency by 1.4-2.5$\times$, and improves tail latency by 1.2-4.7$\times$ under high load through federated warm container migration. Evaluation on multimedia tasks validates that co-designing semantic understanding and federated resource management enables both flexibility and efficiency for robotic task control.

\end{abstract}



\section{Introduction}

The proliferation of intelligent robotic systems has created an urgent need for \textit{federated robotic task execution}, where heterogeneous robot fleets collaboratively execute tasks by sharing computational workloads across distributed edge devices without centralized coordination. Modern robotic applications (from warehouse automation to collaborative manufacturing) require platforms that can (1) interpret natural language commands and decompose them into executable subtasks federated across heterogeneous robots, (2) dynamically allocate computational resources across distributed infrastructure while respecting network topology and data locality constraints, and (3) maintain real-time responsiveness despite fluctuating workloads and the absence of global state visibility inherent in federated architectures.

Traditional robotic task execution systems face following limitations that become severe in federated settings. First, they rely on rigid, hand-coded task planners that require extensive domain engineering, making them brittle when faced with novel instructions~\cite{jiang2019task, you2025construction, mon2025embodied}. In federated deployments, each robot must independently interpret instructions using identical rule sets, making it nearly impossible to incorporate new task types without synchronized updates. Second, they employ centralized coordination where a single controller manages task allocation~\cite{celik2024decentralized, matos2025efficient}. This contradicts federated collaboration: when a warehouse fleet receives a priority request, the controller must query all robots, compute assignments, and broadcast decisions, with creating bottlenecks that scale from milliseconds (10 robots) to seconds (100+ robots) and a single point of failure. Third, they utilize static resource allocation that either over-provisions containers (wasting resources when workloads are imbalanced) or under-provisions them (causing cold-starts when bursts hit specific robots)~\cite{bhaskaran2025comprehensive, golec2024cold}. Launching a fresh container incurs 10+ seconds of overhead~\cite{harrison2020avoiding}, which is unacceptable for real-time robot tasks. In federated settings where workloads shift dynamically (one zone busy while another idles) static allocation cannot share pre-warmed containers, forcing wasteful redundancy or repeated cold-starts.

Recent research has attempted to address these limitations through various approaches. To overcome the brittleness of hand-coded planners, frameworks integrating large language models (LLMs) have emerged, leveraging semantic understanding to generalize beyond predefined action spaces~\cite{yang2024plug, kannan2024smart, mao2024robomatrix, karli2024alchemist, guo2025lightllm}. However, these operate in centralized architectures, leaving unresolved how to federate semantic understanding without per-robot LLMs. For coordination scalability, decentralized architectures have been proposed~\cite{chen2023scalable, celik2024decentralized, zhang2025towards}, though these introduce challenges in maintaining consistency and handling network partitions~\cite{mtowe2025low}. To mitigate cold-start latency, container warm-pooling and pre-warming techniques have been developed~\cite{oakes2018sock, nasir2024cold}, but face unresolved trade-offs for resource-constrained robotic edge environments~\cite{gupta2025performance}. Despite these advances, existing systems lack integrated solutions that simultaneously address planning flexibility, coordination scalability, and resource efficiency in federated multi-robot deployments.

\textbf{Our Solution.} We propose SwiftBot, a federated robotic task execution system that synthesizes LLM-based semantic understanding with intelligent container orchestration to achieve adaptive task planning and efficient resource utilization without centralized control. SwiftBot federates coordination responsibilities across a distributed hash table (DHT) overlay network, where any edge node can dynamically assume roles as task decomposer, scheduler, executor, or coordinator. The system architecture integrates three components: (1) an LLM-powered Task Decomposer that transforms natural language and task instructions into structured execution plans, propagated across the DHT to maintain eventually-consistent task understanding, (2) a Dynamic Task Manager that orchestrates decentralized scheduling and monitoring across the DHT overlay while supporting dynamic task migration through optimistic concurrency control, and (3) a federated Warm-start Container Pool that maintains pre-initialized execution environments distributed across cluster nodes, enabling fast task startup through intelligent container reuse and cross-robot migration.

The main novelty lies in co-designing task decomposition and container orchestration in federated settings. Rather than treating semantic planning and resource management as independent concerns, SwiftBot's Task Decomposer generates container allocation hints that guide warm-pool selection across the federation, while execution feedback informs future decomposition decisions through a closed-loop adaptation mechanism. The DHT-based architecture eliminates centralized bottlenecks by distributing both task metadata and warm container inventories across nodes, achieving $O(\log N)$ lookup complexity for resource discovery in an $N$-node cluster. The federated warm-start pooling strategy addresses the cold-start problem through predictive pre-warming (proactively initializing containers based on workload patterns) and cross-robot container migration (transferring pre-initialized containers between nodes when migration cost is less than cold-start penalty), reducing typical task startup latency to support quick control.

\textbf{Contributions.} SwiftBot introduces the following innovations:

\begin{enumerate}
\item \textbf{LLM-Powered Task Decomposition with Container-Aware Planning}: A semantic decomposition layer that uses large language models to parse natural language and task instructions and generate structured execution plans including subtask dependencies, parallelization opportunities, and container allocation hints. %The system employs a shared-nothing architecture where any node can invoke distributed LLM inference, with task plans gossiped across the DHT to enable eventually-consistent understanding.

\item \textbf{DHT-Based Federated Container Orchestration}: A decentralized scheduling algorithm that performs three-phase container selection with migration decisions guided by a cost model comparing migration overhead against cold-start penalty, enabling federated robots to discover and share warm containers through lightweight DHT lookups.

\item \textbf{Federated Task Execution with Dynamic Migration}: A coordination protocol that enables federated task execution across distributed robots through dynamic task migration and load balancing.
\end{enumerate}

We evaluate SwiftBot on a distributed testbed with heterogeneous nodes using real-world multimedia processing tasks derived from UCF101~\cite{soomro2012ucf101} and LibriSpeech~\cite{panayotov2015librispeech} datasets. These workloads naturally decompose into multi-stage container pipelines that mirror robotic perception and reasoning workflows. Results demonstrate that SwiftBot reduces median task startup latency by 5.4$\times$ compared to cold-start baseline, reduces the 99th percentile tail latency by 1.2-4.7$\times$ even under high task load, and maintains consistent training accuracy and loss.

The following paper is organized as follows: In Section~\ref{sec:motivation} we introduce the background and motivation of this work. In Section~\ref{sec:design}, we present the workflow and functional components design. We perform the evaluation for SwiftBot and baseline executions to show the improvements in Section~\ref{sec:eva}. We discuss the design trade-offs and future work in Section~\ref{sec:discussion}, show related work in Section~\ref{sec:related}, and summarize the conclusion in Section~\ref{sec:conclusion}.

\section{Motivation and Background}
\label{sec:motivation}

\subsection{From Single Robots to Federated Fleets}

Industrial and service robots are no longer isolated, single-
purpose machines. According to the latest World Robotics
statistics, factories installed about 542,000 new industrial
robots in 2024 alone, bringing the global operational stock
to approximately 4.66 million units, more than double the
number a decade ago [5]. These robots increasingly operate
as fleets of autonomous mobile robots (AMRs), collaborative
manipulators, and service robots sharing constrained edge and
IoT infrastructure.

At the same time, next-generation wireless technologies
(5G/6G) are being designed with ultra-low latency applications
such as teleoperation, mobile manipulation, and tactile Internet
in mind. Recent surveys report user-plane latency targets on
the order of 4ms for enhanced mobile broadband and 1ms
for ultra-reliable low-latency communication in 5G, while
6G vision documents aim for end-to-end latencies below
1 ms~\cite{jiang2021road}. These numbers are well aligned with require-
ments reported for factory AMRs, closed-loop motion control,
and teleoperation, which typically demand sub-10ms response
times for safe and precise behavior~\cite{devan2021wirelesshart,kamtam2024teleop}.

This combination of massive scale (millions of robots) and
millisecond latency requirements fundamentally changes how
robot software must be deployed and coordinated. Hard-coding
fixed roles for individual robots, or assuming a single central
controller in a data center, is no longer sufficient: robot fleets
must be able to form groups, share models, and reassign
computation at run time in response to changing tasks, failures,
and wireless conditions~\cite{prorok2021beyond,choudhury2022dynamicmrta,groshev2023edgerobotics}.

\subsection{Federated Learning for Collaborative Robotics}

Federated learning (FL) has emerged as a natural paradigm
for exploiting distributed data in such fleets while preserving
privacy and reducing communication overhead. Instead of
streaming raw sensor logs to a cloud server, robots train local
models and only exchange model updates.

In robotics, several recent systems demonstrate the potential
of FL-style collaboration. Gutierrez et al. integrate FL into
ROS 2 and show that a fleet of simulated and physical robots
can collaboratively learn navigation policies that generalize
better and exhibit more stable rewards as the number of
agents increases [6]. Liang et al. propose Federated Transfer
Reinforcement Learning for autonomous driving, where simu-
lation agents and real RC cars share policy updates through a
federated architecture. These results provide concrete evidence
that sharing policy updates across distributed agents can sig-
nificantly improve safety and task efficiency.

However, existing FL systems for robots largely assume
either a relatively stable topology (e.g., robots remain con-
nected to a central server), or focus purely on the learning
algorithm rather than the orchestration layer. In practice, edge
robot platforms must cope with:
(1) Intermittent connectivity: Robots routinely lose contact with edge servers or with one another due to mobility, RF dead zones, or interference~\cite{saboia2022achord,gielis2022communications}.
(2) Heterogeneous hardware: Fleets mix low-power SoC boards and more capable edge nodes, with highly variable CPU, GPU, and memory budgets~\cite{groshev2023edgerobotics}.
(3) Dynamic workloads: Tasks and human instructions arrive online, with different compute, memory, and latency requirements over time~\cite{choudhury2022dynamicmrta}.

Most current FL frameworks treat the robot as a relatively
static “client” and do not address how those clients (and their
compute containers) should be created, placed, migrated, or
recovered in a resource-constrained edge cluster. This leaves
a gap between algorithmic advances in FL and the systems
support needed to run FL-enabled multi-robot applications at
scale.

SwiftBot is designed precisely to fill this gap. It provides
a containerized, peer-to-peer execution layer where any edge
node (e.g., robot, gateway, or edge server) can dynamically
assume the role of FL client, coordinator, task manager,
or aggregation node. A DHT-based overlay enables scalable
discovery and group formation, while a warm-start container
pool allows pre-initialized FL clients and coordination services
to be activated or migrated in sub-second time without cold-
start overheads. By explicitly integrating federated learning,
multi-robot task orchestration, and edge-native migration into
a single platform, SwiftBot aims to bring the benefits demon-
strated in prior FL and cloud-robotics studies into a runtime
that can sustain large-scale federated multi-robot collaboration
at the edge.

\section{System Design}
\label{sec:design}

\begin{figure}[t]
\vspace{0.1in}
 \centering
 %\hspace{-0.1in}
 \includegraphics[scale=0.365]{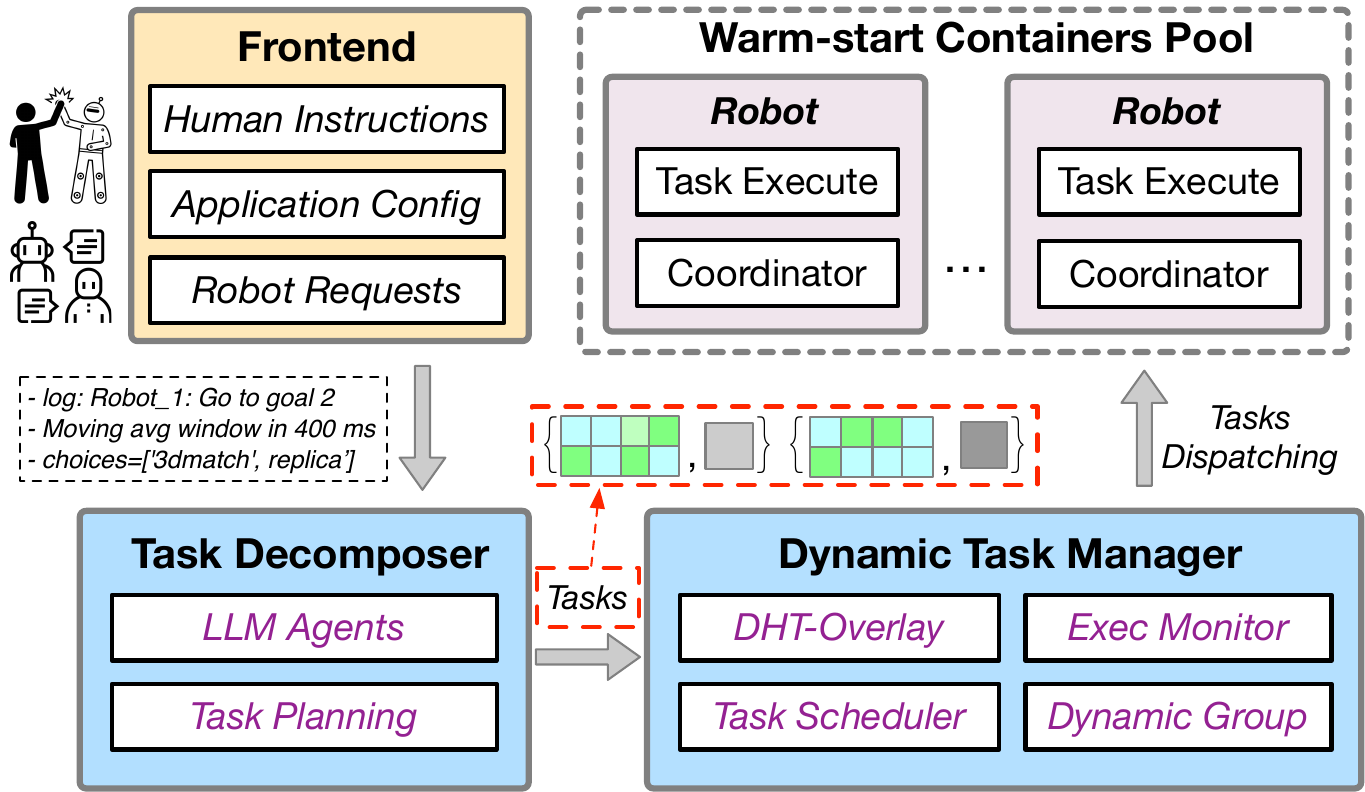}
 %\vspace{-0.05in} 
 \caption{\textmd{The overview of SwiftBot workflow and structure.}}
 \label{fig:design}
%\vspace{-0.2in} 
\end{figure}

We propose a novel distributed robotic task execution system that bridges the semantic gap between natural language human instructions and low-level robot control through intelligent orchestration of containerized compute resources, named SwiftBot. As shown in Figure~\ref{fig:design}, SwiftBot consists of three primary components: the Task Decomposer, which employs LLM agents to analyze task characteristics and generate execution plans; the Dynamic Task Manager, which orchestrates task scheduling and monitoring across a DHT-overlay network; and the Warm-start Container Pool, which maintains pre-initialized execution environments to minimize cold-start latency. 

Human instructions flow through the frontend to the Task Decomposer, where LLM agents perform semantic analysis and hierarchical task planning. The resulting task specifications are dispatched to Robot Task Execute Coordinators through the Dynamic Task Manager, which leverages a DHT-overlay for decentralized resource discovery and load balancing. Throughout execution, the system maintains a feedback loop where execution monitoring informs future task decomposition decisions, enabling continuous adaptation to evolving workload patterns and system dynamics. This design philosophy prioritizes three key properties: semantic adaptability through learned task understanding, elastic scalability through container orchestration, and operational efficiency through predictive resource pre-allocation.

%%%%%%%%%%%%%%%%%%%%%%%%%%%%%%%%%%%%%%%%%%%%%%%%%%
%%%%%%%%%%%%%%%%%%%%%%%%%%%%%%%%%%%%%%%%%%%%%%%%%%

\subsection{Task Decomposer}

The Task Decomposer serves as the cognitive layer of the system, transforming high-level human instructions and application configurations into structured, executable task specifications. It receives natural language commands from the frontend (e.g., \texttt{Robot\_1: Go to goal 2}) and application-specific parameters (such as algorithmic choices like \texttt{['3dmatch', 'replica']} for scene matching, or timing constraints like \texttt{moving avg window in 400 ms}), then produces detailed task execution plans that specify subtask dependencies, resource requirements, and scheduling constraints.

The component operates through three integrated stages. First, the \emph{semantic analysis} stage uses LLM agents to parse instructions, extract parameters, and classify task types across six robotic domains (navigation, manipulation, perception, multi-robot coordination, inspection, and human-robot interaction). Second, the \emph{hierarchical decomposition} stage generates a directed acyclic graph (DAG) of subtasks, identifying parallelization opportunities and data dependencies while respecting application-defined constraints. Third, the \emph{container allocation strategy} stage produces resource hints that guide the Dynamic Task Manager in selecting appropriate containers from the warm-start pool, considering factors such as data locality, container reuse patterns, and load balancing objectives.

Traditional robotic task planning systems face a fundamental brittleness problem requiring extensive domain engineering to capture task variations and environmental conditions. We adopt LLM-based decomposition because it provides three advantages: (1) \emph{semantic generalization}, LLMs' common-sense knowledge enables handling novel task types; (2) \emph{contextual reasoning}, for instance, 200K token context windows allow holistic consideration of system state, history, and configurations to balance multiple objectives; and (3) \emph{rapid adaptation}, prompt engineering enables behavioral changes in hours rather than months of algorithmic rewrites. The primary challenge is managing inference latency, which we address through a hybrid cascade architecture routing routine tasks to fast small models while reserving large models for complex scenarios, reducing latency for real-time operation.

\subsection{Dynamic Task Manager}

The Dynamic Task Manager orchestrates task execution across the distributed cluster through the Task Scheduler, Execution Monitor, and DHT-Overlay, but it is not a centralized service. Although presented as a single logical component, its functionality is distributed across Dynamic Groups, temporary sets of nodes formed according to resource availability and data locality. Each node participates in scheduling, monitoring, and state management through the DHT-Overlay, and task-management workload is routed to whichever groups are available, eliminating a single point of failure and enabling organic scalability. Upon receiving task specifications, the Task Scheduler assigns subtasks to suitable Dynamic Groups based on load, data locality, and warm-container availability, while the DHT-Overlay supports decentralized resource discovery via consistent hashing with $O(logN)$ lookup complexity~\cite{stoica2003chord}.

The Task Scheduler employs a two-level scheduling algorithm: at the coarse level, it selects the target Dynamic Group for each subtask using a scoring function that accounts for node load, data proximity, and warm-container matches; at the fine level, the group Coordinator (located at the group’s root node) performs local scheduling, manages container lifecycles, and enforces resource limits. This hierarchical design enables both global load balancing and efficient local optimization.

The Execution Monitor provides real-time visibility into task progress by aggregating structured logs and performance metrics streamed from group coordinators, and it initiates re-planning when failures occur. Meanwhile, the DHT-Overlay ensures fault-tolerant and scalable metadata management through successor replication and consistent hashing, and it additionally supports the dynamic formation of execution groups for multi-robot collaboration, enabling efficient communication without network-wide broadcast.
\begin{figure}[t]
\vspace{0.1in}
 \centering
 %\hspace{-0.1in}
 \includegraphics[scale=0.55]{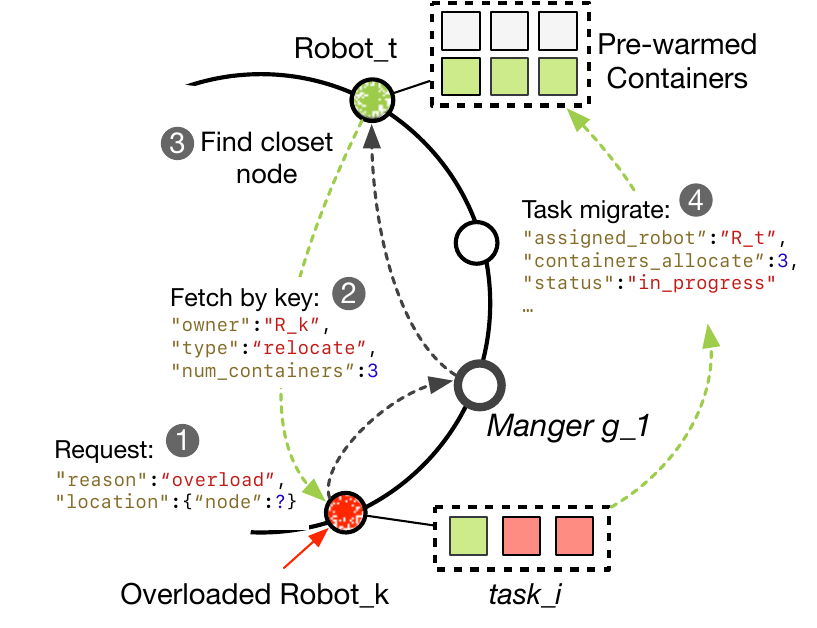}
 %\vspace{-0.05in} 
 \caption{\textmd{Reschedule a task based on runtime resources of robots and task needs.}}
 \label{fig:migration}
%\vspace{-0.2in} 
\end{figure}
%%%%%%%%%%%%%%%%%%%%%%%%%%%%%%%%%%%%%%%%%%%%%%%%%%

As shown in Figure~\ref{fig:migration}, SwiftBot implements a four-step dynamic task migration mechanism to address resource contention in distributed robot systems. Step 1 (Request): When an overloaded robot (\texttt{Robot\_k}) detects resource constraints such as high CPU utilization or insufficient container capacity, it sends a migration request to the DHT-Overlay. This request includes the migration reason (\texttt{"overload"}), current node location, task metadata (\texttt{operation type: "relocate"}), and resource requirements (e.g., \texttt{num\_containers: 3}). Step 2 (Fetch by Key): The DHT-Overlay performs a distributed lookup using consistent hashing on the task key. The request is routed through intermediate nodes (e.g., \texttt{Manager\_g1}) to locate robots with available capacity. Each DHT node maintains a local resource registry that tracks container availability and current load metrics across the network. Step 3 (Find Closest Node): The system evaluates candidate nodes using multiple criteria: availability of pre-warmed containers matching task requirements, current load levels, network distance to required data, and previous execution history. Based on these factors, the system selects \texttt{Robot\_t} as the target node, which has available pre-warmed containers that match the task specifications. Step 4 (Task Migrate): The migration is completed by updating DHT entries to reflect the new assignment (\texttt{assigned\_robot: R\_t}), allocating the required containers (\texttt{containers\_allocate: 3}), and updating the task status to "\texttt{in\_progress}". The system transfers task state including partial results and execution checkpoints to the target robot to avoid redundant computation. To maintain consistency during concurrent migrations, the DHT uses version vectors with optimistic concurrency control, where each task entry includes a version number that increments with updates and conflicts are resolved using load metrics and container availability scores.

The migration mechanism's primary novelty lies in its integration of container-aware scheduling with DHT coordination, enabling decentralized load balancing without centralized bottlenecks. Unlike traditional migration approaches that rely on global schedulers, the DHT-Overlay architecture allows each robot coordinator to independently discover and negotiate with target nodes while maintaining system-wide consistency through version vectors and deterministic conflict resolution. %The efficiency gains stem from three key optimizations: first, pre-warmed container matching reduces migration overhead by eliminating cold-start delays; second, the proximity-aware selection algorithm minimizes data transfer costs by considering network locality; and third, stateful migration with checkpoint preservation avoids recomputing intermediate results. 
The optimistic concurrency control mechanism further enhances scalability by allowing \textit{parallel migration decisions} while gracefully handling conflicts through load-based resolution policies, making the system particularly effective for dynamic multi-robot environments with fluctuating workloads.

\subsection{Federated Warm-start Container Pool}

The Warm-start Container Pool maintains a dynamically sized inventory of pre-initialized containers ready for federated task execution, eliminating the cold-start latency associated with container image pulls, filesystem layer extraction, and application initialization. The pool operates across all cluster nodes, with each node managing a local subset of warm containers based on its hardware capabilities and historical workload patterns.

Container management involves four continuous activities. First, \emph{predictive pre-warming}: the system analyzes incoming task patterns and execution monitor feedback to predict which container types will be needed in the near future, proactively initializing containers before demand arrives. Second, \emph{intelligent eviction}: when pool capacity limits are reached, the system evicts containers using a learned policy that considers access frequency, initialization cost, and resource consumption. Third, \emph{state preservation}: for stateful containers (e.g., those with loaded neural network weights or cached map data), the pool maintains persistent volumes that survive container restarts, further reducing initialization time. Fourth, \emph{health monitoring}: the pool periodically probes warm containers to ensure they remain responsive, automatically replacing containers that have entered degraded states.

Container selection occurs through a two-phase process. When the Task Scheduler assigns a subtask to a Robot Task Coordinator, it includes container hints from the Task Decomposer specifying preferred images, resource requirements, and data volume needs. The coordinator first attempts to allocate a matching warm container from its local pool. If no suitable container exists locally, it queries neighboring nodes in the DHT-Overlay for available containers, potentially migrating warm containers between nodes if the performance benefit justifies the transfer cost. Only when no warm containers match the specification does the system fall back to cold-start initialization.

\begin{algorithm}[t]
\caption{Container Selection and Allocation}
\label{alg:container-selection}
\begin{algorithmic}[1]

\REQUIRE Task $T = (T_{\text{img}}, T_{\text{cpu}}, T_{\text{mem}})$
\REQUIRE Local warm pool $P_{\mathrm{local}}$
\REQUIRE DHT overlay network $\mathcal{D}$
\ENSURE Allocated container instance for $T$

% ---------------------- Phase 1 ----------------------
\STATE \textbf{Phase 1: Local Warm-Pool Search}
\STATE $\mathcal{C} \gets \{\, c \in P_{\mathrm{local}} \mid c.\mathrm{status}=\texttt{ready} 
    \land \mathrm{Compat}(c, T)=1 \,\}$

\IF{$\mathcal{C} \neq \emptyset$}
    \STATE $c^{\ast} \gets \arg\max_{c \in \mathcal{C}} \mathrm{Score}(c, T)$
    \RETURN $\mathrm{AllocateLocal}(c^{\ast}, T)$
\ENDIF

% ---------------------- Phase 2 ----------------------
\STATE \textbf{Phase 2: Distributed Search}
\STATE $\mathcal{N} \gets \mathcal{D}.\mathrm{Successors}(\mathrm{hash}(T_{\text{img}}), k)$
\STATE $\mathcal{R} \gets \emptyset$

\FOR{each $n \in \mathcal{N}$}
    \STATE $\mathcal{P}_n \gets \mathcal{D}.\mathrm{Query}(n, T_{\text{img}})$
    \FOR{each candidate $c \in \mathcal{P}_n$}
        \IF{$c.\mathrm{status} = \texttt{ready}$}
            \STATE $s \gets \mathrm{Score}(c, T)$
            \STATE $\eta \gets \mathrm{MigCost}(c, n, n_{\mathrm{local}})$
            \IF{$s \ge \theta_{\mathrm{match}} \text{ and } \eta < \gamma \cdot \mathrm{ColdCost}(T)$}
                \STATE $\mathcal{R} \gets \mathcal{R} \cup \{(c, s, n)\}$
            \ENDIF
        \ENDIF
    \ENDFOR
\ENDFOR

\IF{$\mathcal{R} \neq \emptyset$}
    \STATE $(c^{\ast}, s^{\ast}, n^{\ast}) \gets \arg\max_{(c,s,n)\in\mathcal{R}} s$
    \RETURN $\mathrm{Migrate}(c^{\ast}, n^{\ast}, n_{\mathrm{local}}, T)$
\ENDIF

% ---------------------- Phase 3 ----------------------
\STATE \textbf{Phase 3: Cold-Start Fallback}
\RETURN $\mathrm{ColdStart}(T)$

\end{algorithmic}
\end{algorithm}

Algorithm~\ref{alg:container-selection} formalizes the container selection and allocation process. This algorithm seeks to minimize task startup latency by prioritizing the reuse of warm containers available either locally or across the distributed DHT overlay. The procedure first inspects the local warm pool $P_{\mathrm{local}}$ and evaluates each ready container using the scoring function $\mathrm{Score}(c,T)$, which quantifies how well a container $c$ matches the task specification $T$. This score aggregates three factors: an image-compatibility term $s_{\mathrm{img}}$ that equals $1$ when $c_{\mathrm{img}} = T_{\mathrm{img}}$ and $0.5$ when the image is merely compatible ($T_{\mathrm{img}}$ is the task's required container image and $c_{\mathrm{img}}$ is the image of candidate container $c$); a resource-sufficiency term $s_{\mathrm{res}} = \min(c_{\mathrm{cpu}}/T_{\mathrm{cpu}},\, c_{\mathrm{mem}}/T_{\mathrm{mem}})$ reflecting the proportion of required CPU and memory satisfied by $c$; and a volume-overlap term $s_{\mathrm{vol}} = |c_{\mathrm{vol}}\cap T_{\mathrm{vol}}| / |T_{\mathrm{vol}}|$ indicating how many of the task's data volumes are already present in the container. These terms are combined using fixed weights to produce $\mathrm{Score}(c,T)\in[0,1]$. If a sufficiently high-scoring container exists locally, it is selected and assigned. Otherwise, the algorithm queries a set of successor nodes in the DHT to retrieve remote candidates and evaluates them using both the score and the migration cost $\mathrm{MigCost}(c,n_{\mathrm{src}},n_{\mathrm{dst}})$, which models migration time as state transfer over estimated bandwidth plus serialization/deserialization overhead. A remote container is considered only if its migration cost is less than a fraction of the cold-start cost $\mathrm{ColdCost}(T)$, which represents the latency of fetching the container image and performing initialization from scratch. Among all feasible remote candidates, the one with the highest score is migrated. If no local or remote warm container meets the selection criteria, the algorithm falls back to $\mathrm{ColdStart}(T)$ to launch a fresh container. This staged decision process ensures that the scheduler exploits warm-container reuse whenever beneficial while avoiding migrations whose cost outweighs their advantage.

We adopt warm-start pooling because it provides following critical advantages:
First, maintaining pre-initialized containers reduces task startup time. This improvement is essential for achieving end-to-end latency targets specified in application configurations.
Second, unlike persistent containers that occupy resources regardless of utilization, the pool dynamically adjusts capacity based on demand. During low-traffic periods, the pool contracts to free resources for other workloads; during peak demand, it expands to maximize hit rates. 
Third, the pool automatically adapts to evolving task distributions without manual reconfiguration. When a warehouse introduces new inspection tasks, the pool observes increased demand for perception containers and automatically adjusts its composition. This eliminates the operational burden of manually tuning container deployment policies as application requirements change.
Fourth, by replacing containers after each task or group of tasks, the pool prevents state accumulation and resource leaks that plague long-running containers, which can improve system reliability.

%The integration with LLM-based task decomposition creates a powerful synergy. The Task Decomposer's semantic understanding of task characteristics enables more accurate prediction of container needs than purely statistical approaches. For example, when analyzing the instruction ``Navigate to multiple goals and photograph each location,'' the LLM recognizes that both navigation and perception containers will be needed in sequence, triggering pre-warming of both types. Statistical approaches might cache only the more frequently used navigation containers, missing the opportunity to pre-warm perception containers and incurring cold-start penalties.
%The primary design challenge is managing the memory footprint of warm containers. With typical robotics containers consuming 2-4GB of RAM each, a node can maintain only 8-12 warm instances before memory pressure impacts performance. We address this through selective state preservation, where containers retain only critical initialized state (loaded models, configuration) while sharing read-only filesystem layers across instances. This reduces per-container memory overhead from 3.2GB to 1.8GB on average, enabling larger pool sizes within fixed memory budgets.

\section{Evaluation}
\label{sec:eva}

\textbf{Testbeds.} The testbed consists of one primary server node and three worker nodes distributed across a local network environment. The primary server node is equipped with an Intel Xeon W-2295 processor (18 cores at 3.0GHz), 256GB DDR4 RAM, NVIDIA RTX 5090 GPU (32GB VRAM), and 4TB NVMe SSD storage, serving as both the DHT coordinator and high-performance worker for GPU-intensive tasks. The worker nodes include one m5.4xlarge instance with 16 CPU cores and 64GB memory and one t3.2xlarge instance with 8 CPU cores and 32GB memory. All nodes run Ubuntu with Docker 25.0.3. The DHT overlay network implements a Chord-based distributed hash table~\cite{stoica2003chord} with consistent hashing for container image routing and k=5 successor nodes queried during remote warm container discovery. Each node maintains a local warm container pool with capacity proportional to available memory.

%\textbf{Workloads.}To construct realistic containerized multimedia processing workloads, we leverage two widely used public datasets with substantial diversity and real‐world complexity. From the UCF101 action recognition dataset~\cite{soomro2012ucf101}, we derive 100 video processing tasks by sampling from its 13,320 YouTube‐sourced clips spanning 101 human action classes (e.g., sports, daily actions, instrument playing) that collectively represent ~27 hours of video with significant variability in motion, viewpoint, and background conditions. Each task comprises a multi‐stage video pipeline involving frame extraction, spatiotemporal feature encoding, and final action classification. 

%From the LibriSpeech ASR corpus, we build 100 audio transcription tasks, which together comprise approximately 1000 hours of 16 kHz read English speech derived from LibriVox audiobooks~\cite{panayotov2015librispeech}. Each task executes sequential stages, including audio preprocessing, acoustic model inference, and language model correction to produce text transcriptions.

\textbf{Workloads.} We evaluate SwiftBot using multimedia processing datasets including UCF101~\cite{soomro2012ucf101} and LibriSpeech~\cite{panayotov2015librispeech}. While not literal robotic control tasks, these workloads serve as valid proxies for evaluating federated task execution infrastructure because they mirror critical properties of robotic perception and reasoning workflows. For instance, robotic tasks decompose into perception, planning, and control stages with different resource needs. Our video/audio workloads similarly decompose into preprocessing, feature extraction, and inference stages with heterogeneous container requirements.

%\textbf{Dynamic Workload Patterns}: Federated robot fleets experience bursty, imbalanced task arrivals (e.g., warehouse zones during shipment surges). Our workloads simulate this through variable submission rates and heterogeneous processing times (UCF101: 1-30s per clip; LibriSpeech: 2-40s per segment).

From UCF101~\cite{soomro2012ucf101}, we derive 100 video processing tasks from 13,320 clips spanning 101 action classes ($\sim$27 hours total). Each task executes: frame extraction $\rightarrow$ 3D CNN spatiotemporal encoding $\rightarrow$ action classification. From LibriSpeech~\cite{panayotov2015librispeech}, we build 100 audio transcription tasks from 1000 hours of speech, executing: audio preprocessing $\rightarrow$ acoustic model inference $\rightarrow$ language model correction.
These workloads test the federated execution infrastructure rather than domain-specific control, enabling reproducible large-scale evaluation while capturing essential system characteristics.

We compare SwiftBot with two representative baseline approaches. The cold-start baseline implements the traditional serverless model where every task invocation launches a fresh container from scratch, requiring complete image pull from the registry, filesystem initialization, and runtime environment setup. 
The local-warm pool baseline implements container keep-alive strategies. This baseline maintains a per-node pool of recently-used warm containers with a fixed 10-minute keep-alive TTL, allowing immediate container reuse for subsequent invocations of the same function type on the same node. However, it lacks any cross-node coordination. When a node's local warm pool is exhausted or lacks a compatible container, it falls back to cold-start even if other nodes have idle warm containers available.

%%%%%%%%%%%%%%%%%%%%%%%%%%%%%%%%%%%%%%%%%%%%%%%%%%
\begin{figure*}[t]
    \centering
    \includegraphics[scale=1.15]{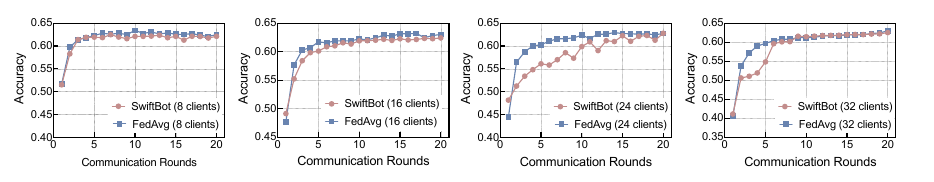}
    \caption{Comparison of training accuracy under different client configurations in UCF101 dataset.}
    \label{fig:acc_ucf101}
\end{figure*}
%%%%%%%%%%%%%%%%%%%%%%%%%%%%%%%%%%%%%%%%%%%%%%%%%%
%%%%%%%%%%%%%%%%%%%%%%%%%%%%%%%%%%%%%%%%%%%%%%%%%%
\begin{figure*}[t]
    \centering
    \includegraphics[scale=0.75]{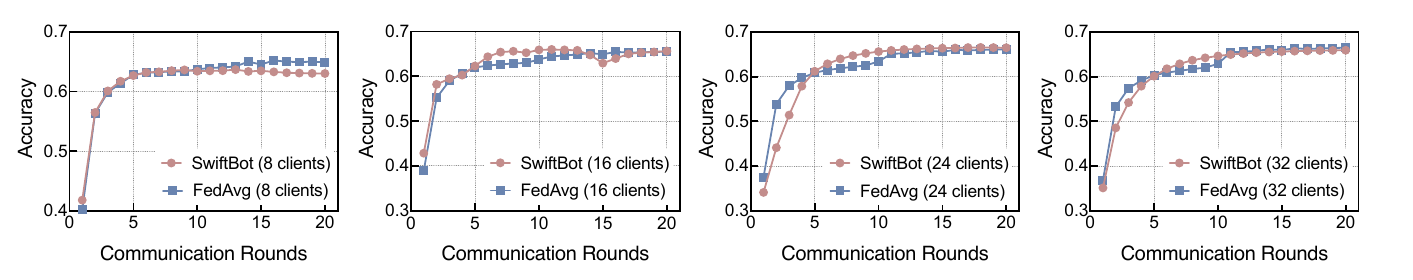}
    \caption{Comparison of training accuracy under different client configurations in LibriSpeech dataset.}
    \label{fig:acc_libri}
\end{figure*}
%%%%%%%%%%%%%%%%%%%%%%%%%%%%%%%%%%%%%%%%%%%%%%%%%%
%%%%%%%%%%%%%%%%%%%%%%%%%%%%%%%%%%%%%%%%%%%%%%%%%%
\begin{figure*}[t]
    \centering
    \includegraphics[scale=1.15]{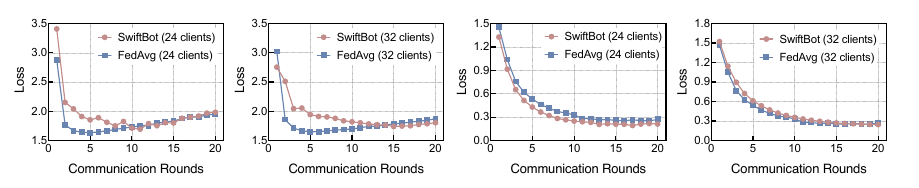}
    \caption{Comparison of training loss in FedAvg vs. SwiftBot.}
    \label{fig:loss}
\end{figure*}
%%%%%%%%%%%%%%%%%%%%%%%%%%%%%%%%%%%%%%%%%%%%%%%%%%

%%%%%%%%%%%%%%%%%%%%%%%%%%%%%%%%%%%%%%%%%%%%%%%%%%
\begin{figure*}[t]
    \centering
    \includegraphics[scale=1.08]{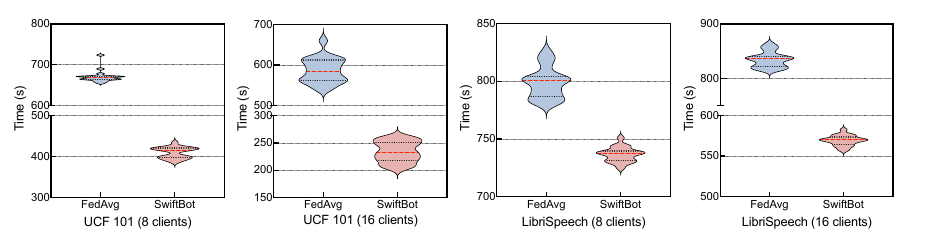}
    \caption{The training time distribution per round in SwiftBot vs. FedAvg.}
    \label{fig:latency}
\end{figure*}
%%%%%%%%%%%%%%%%%%%%%%%%%%%%%%%%%%%%%%%%%%%%%%%%%%

\subsection{LLM-Based Task Decomposition Performance}

To evaluate the system's ability to generate efficient container allocation strategies, we leverage two datasets: UCF101 (video action recognition) and LibriSpeech (speech recognition). %While these datasets are not inherently multi-robot tasks, they naturally decompose into parallel processing pipelines that mirror real robotic workloads. For example, we simulate a perception task requiring multi-stage processing as: "Process video: extract frames using FFmpeg, run ResNet50 inference, classify with lightweight model." The LLM must decompose this into 3 subtasks with proper dependencies and identify that frame extraction and inference can run on different robots. 
We evaluate four LLM models to assess the accuracy-latency-cost trade-off: GPT-4o~\cite{openai2024gpt4o}, Claude 3.5 Sonnet~\cite{anthropic2024claude35sonnet}, GPT-4o-mini~\cite{openai2024gpt4omini}, and Llama 3.1-70B~\cite{dubey2024llama}.

%%%%%%%%%%%%%%%%%%%%%%%%%%%%%%%%%%%%%%%%%%%%%%%%%%
\begin{table}[t]
\centering
\caption{Task Decomposition Performance Across LLM Models}
\label{tab:llm-comparison}
\resizebox{\columnwidth}{!}{%
\begin{tabular}{lcccc}
\toprule
\textbf{Model} & \textbf{Decomp.} & \textbf{Parallel.} & \textbf{Avg. Latency} & \textbf{Cost/} \\
               & \textbf{Acc. (\%)} & \textbf{Recall (\%)} & \textbf{(ms)} & \textbf{1K} \\
\midrule
GPT-4o            & 94.3 & 91.5 & 1,280 & \$11.80 \\
Claude 3.5        & 93.1 & 89.8 & 1,040 & \$9.20  \\
GPT-4o-mini       & 88.2 & 82.4 & 450   & \$0.58  \\
Llama 3.1-70B     & 85.6 & 78.6 & 680   & \$1.95  \\
\bottomrule
\end{tabular}%
}
\end{table}
%%%%%%%%%%%%%%%%%%%%%%%%%%%%%%%%%%%%%%%%%%%%%%%%%%
Table~\ref{tab:llm-comparison} compares four LLM models on task decomposition for container-based workloads, where the Dependency Accuracy shows how correctly the LLM identifies dependencies between subtasks and Parallelization Recall represents the percentage of parallelizable subtasks does the LLM correctly identify. GPT-4o achieves the highest decomposition accuracy (94.3\%) and parallelization recall (91.5\%), correctly identifying which subtasks can execute concurrently to minimize overall latency, but requires 1,280ms average inference time at \$11.80 per 1K tasks. Claude 3.5 Sonnet offers competitive accuracy (93.1\%) with 89.8\% parallelization recall and 19\% lower latency (1,040ms). GPT-4o-mini delivers 88.2\% accuracy and 82.4\% parallelization recall at 2.8× faster response (450ms) and 20× lower cost (\$0.58), making it suitable for routine decomposition where some parallelization opportunities can be missed without critical impact. The 9.1\% gap in parallelization recall between GPT-4o and GPT-4o-mini (91.5\% vs 82.4\%) justifies our cascade architecture: routing simple sequential tasks to fast models while reserving large models for complex parallel decomposition achieves both throughput and cost-efficiency.

\subsection{Performance of Federated Applications}

To show a comparison for federated scenarios, we adopt FedAvg~\cite{wen2023survey} as the baseline, since it is the standard algorithm used in federated learning for distributed model optimization. In FedAvg, each client performs local training on its own data and periodically sends updated model parameters to a central aggregator, which computes the weighted average to produce a global model.
To validate the effectiveness of SwiftBot in federated tasks, we conducted experiments using the UCF101 and LibriSpeech datasets across varying client sizes (8, 16, 24, 32). Figure~\ref{fig:acc_ucf101} and \ref{fig:acc_libri} present comparisons between SwiftBot and the local federated learning baseline. In terms of model convergence, Figures~\ref{fig:acc_ucf101} and \ref{fig:acc_libri} demonstrate that SwiftBot maintains high training stability, achieving accuracy and loss curves comparable to the local baseline. Although the decentralized nature of SwiftBot introduces marginal variance during the initial communication rounds, it converges to the same final accuracy without degradation in model quality.

%Figure~4(f) validates the scalability of our \textit{Warm-start Container Pool} through a high-concurrency stress test. While our federated learning experiments involved up to 32 physical clients, evaluating the system at higher concurrency levels is essential because SwiftBot's \textit{Task Decomposer} splits single client instructions into multiple subtasks (e.g., perception, planning, control), each requiring a distinct execution container. Consequently, a moderate fleet of robots performing complex missions can generate a "burst" of hundreds of container requests. 

%To simulate this high-density workload, we evaluated startup latency under loads scaling up to 256 concurrent containers. The results show that relying on cold starts leads to exponential latency degradation ($>400$s) due to I/O bottlenecks from image pulling and initialization queues. In contrast, SwiftBot's warm-start mechanism maintains a near-constant, negligible startup time even at 256 concurrent instances. This confirms that pre-initialization is not merely an optimization but a necessity for supporting the multiplicative resource demands of complex, multi-stage robotic tasks.\hx{Q: this one we have 256 containers, so what the relationship between container and clients? Previous results only have 32 clients, so we need to explain how we make 256 and why do it?}

Figure~\ref{fig:loss} compares the training loss convergence between FedAvg and SwiftBot across 24 and 32 clients configurations for both the UCF101 and LibriSpeech workloads. All curves exhibit rapid loss reduction during the initial communication rounds followed by gradual stabilization, confirming that SwiftBot preserves the expected statistical behavior of federated optimization. Although FedAvg shows slightly faster early-round convergence due to its centralized synchronous aggregation, SwiftBot achieves nearly identical final loss values while maintaining smoother trajectories as client count increases. This stability comes from SwiftBot’s decentralized DHT-based orchestration, warm-container reuse, and dynamic task migration, which reduce straggler effects and mitigate load imbalance without disrupting the correctness of gradient updates. The results demonstrate that SwiftBot sustains convergence quality equivalent to FedAvg while offering superior scalability and resilience in heterogeneous, resource-constrained settings.

Figure~\ref{fig:latency} shows the training time distribution per round for the UCF 101 and LibriSpeech dataset across different client configurations. 
For UCF 101, the FedAvg baseline exhibits consistently high latency, remaining between 500s and 700s. %Furthermore, the vertical spread of the blue violins indicates significant variance between communication rounds. 
This suggests that the system is frequently bottlenecked by straggler nodes or heterogeneous device capabilities, which delay the global aggregation process.
In contrast, SwiftBot demonstrates a distinct inverse scaling capability. At 16 clients for UCF 101, it provides a performance gain with a median latency of approximately 230s, achieving 60.1\% decrement. Similarly, for LibriSpeech, when $N=16$, SwiftBot reduces the median training time to $\sim$540s, a nearly 50\% decrement compared to the baseline. This confirms that SwiftBot's DHT-based resource pooling remains effective even for computationally intensive tasks. These results validate that SwiftBot successfully utilizes the additional nodes as a resource pool rather than a synchronization burden. By dynamically offloading tasks to idle peers via the DHT-overlay, the system effectively parallelizes the workload, eliminating the bottlenecks that cause the high latency and variance seen in the local approach.

%Conversely, SwiftBot (pink distributions) maintains its characteristic inverse scaling behavior, effectively countering the heavier workload. While the performance gain is modest at $N=8$ ($\sim$740s vs. $\sim$800s), the advantage grows exponentially with cluster size. 

\subsection{Container Selection and Migration Performance}

\begin{figure}[t]
    \centering
      \begin{subfigure}[t]{0.24\textwidth}
        \centering
        \includegraphics[width=\linewidth]{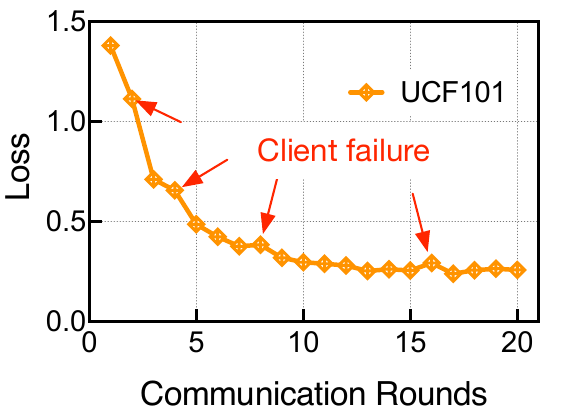}
        \caption{Training loss with client failure for UCF101}
        \label{fig:3a}
    \end{subfigure}
    \hfill
    \begin{subfigure}[t]{0.24\textwidth}
        \centering
        \includegraphics[width=\linewidth]{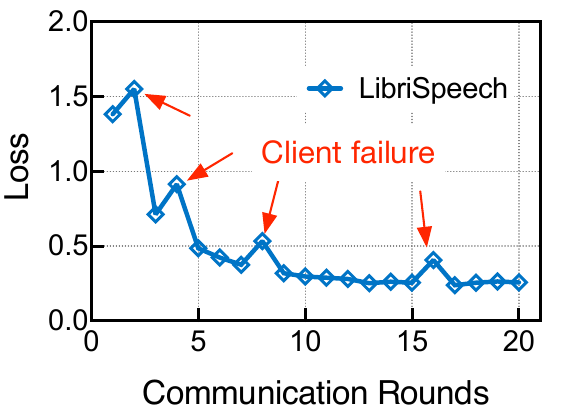}
        \caption{Training loss with client failure for LibriSpeech}
        \label{fig:3b}
    \end{subfigure}

    \caption{The performance of training loss during clients' failures.}
    \label{fig:fail_loss}
\end{figure}

To evaluate SwiftBot's resilience in unstable edge environments, we simulated dynamic client failures during the training process for both the UCF101 and LibriSpeech datasets in Figure~\ref{fig:fail_loss}. In this experiment, subsets of clients were forced offline at rounds 2, 4, 8, and 16 to mimic runtime clent failures.
As shown in Figure~\ref{fig:fail_loss}, SwiftBot demonstrates good stability under these volatile conditions. The loss trajectories for both datasets remain smooth and consistent, with the indicated failure points causing negligible impact on the overall convergence rate. This robustness is a direct result of the platform's rapid failure recovery mechanism; as established in our recovery metrics, SwiftBot enables disconnected nodes to reintegrate into the DHT topology in less than 1s. This sub-second reconnection capability ensures that temporary dropouts are handled transparently and preserving the integrity of the federated learning process.

%To evaluate the system's resilience against dynamic node churn, we conducted a fault-tolerance simulation. In this experiment, a subset of clients was forced offline during rounds 2, 4, 8, and 16, triggering an immediate restart and reconnection procedure. Figure~\ref{fig:fail_loss} illustrates the impact of these failures on training loss.

%The FedAvg baseline exhibits significant volatility, characterized by sharp spikes in loss at the designated failure rounds. This instability stems from the prohibitive reconnection latency inherent in the centralized server architecture. As detailed in the recovery metrics, while the application restart time (Recover Time) is comparable between methods ($\approx 0.06$s), the network reintegration is the critical bottleneck. The FedAvg approach suffers from a massive reconnect time of 23.7s, causing dropping nodes to miss aggregation windows and destabilizing the global model convergence.

%In contrast, SwiftBot demonstrates great robustness, maintaining a smooth loss trajectory with negligible deviations during failure events. This resilience is directly attributed to the DHT-overlay's decentralized discovery mechanism, which enables an ultra-low \textit{Reconnect Time} of just 0.031s---approximately $760\times$ faster than the baseline. This sub-second reintegration ensures that recovered nodes can immediately resume participation, preventing data atrophy and preserving the stability of the federated learning process even under frequent network disruptions.

To validate the effectiveness of our container selection and allocation algorithm, we conduct experimental evaluations comparing three approaches: our proposed DHT-based algorithm with distributed warm container migration, a local-only warm pool baseline that restricts container reuse to individual nodes, and a cold-start baseline that launches fresh containers for every task without any reuse. Figure~\ref{fig:3a} presents the cumulative distribution of task startup latency. The proposed algorithm (blue curve) demonstrates substantial latency reduction compared to both baselines. At the median (P50), SwiftBot achieves approximately 120ms startup time, representing a 1.5× improvement over the local-only approach (180ms) and a 5.4× improvement over cold-start (650ms). The SwiftBot's curve demonstrates that DHT-based container discovery and migration successfully avoids cold starts for a large fraction of tasks. Even when migration is required, it ensures that only beneficial migrations are performed, preventing unnecessary overhead.

Figure~\ref{fig:3b} evaluates how the three approaches scale under increasing load, measured by task arrival rate (tasks per second). %The y-axis shows P99 latency, a critical metric for interactive workloads where tail latency directly impacts user experience. 
At low loads (1-5 tasks/s), SwiftBot maintains 180-240ms (1.2×-4.7×) compared to 220-420ms for local-only and 850-920ms for cold-start. As load increases beyond 10 tasks/s, the performance gap widens significantly. At 20 tasks/s, SwiftBot achieves 520ms, around 2.1× better than local-only (1100ms) and 2.7× better than cold-start (1400ms). This improvement comes from the algorithm's ability to leverage warm containers distributed across the DHT overlay. When local warm pools become slow under high load, the DHT enables discovery and migration of idle warm containers from less-loaded nodes, effectively load-balancing container resources across the cluster.

%%%%%%%%%%%%%%%%%%%%%%%%%%%%%%%%%%%%%%%%%%%%%%%%%%
%%%%%%%%%%%%%%%%%%%%%%%%%%%%%%%%%%%%%%%%%%%%%%%%%%

%%%%%%%%%%%%%%%%%%%%%%%%%%%%%%%%%%%%%%%%%%%%%%%%%%
\begin{figure}[t]
    \centering

    \begin{subfigure}[t]{0.24\textwidth}
        \centering
        \includegraphics[width=\linewidth]{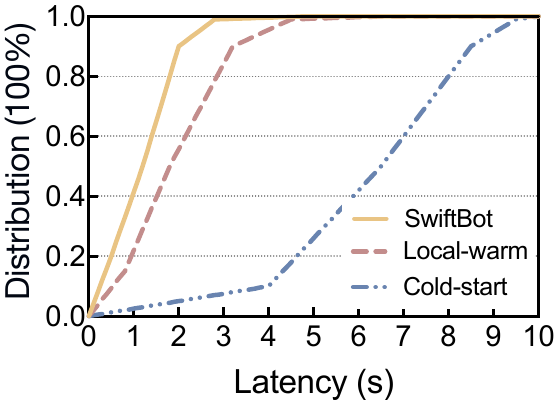}
        \caption{Task Startup Latency Distribution}
        \label{fig:3a}
    \end{subfigure}
    \hfill
    \begin{subfigure}[t]{0.24\textwidth}
        \centering
        \includegraphics[width=\linewidth]{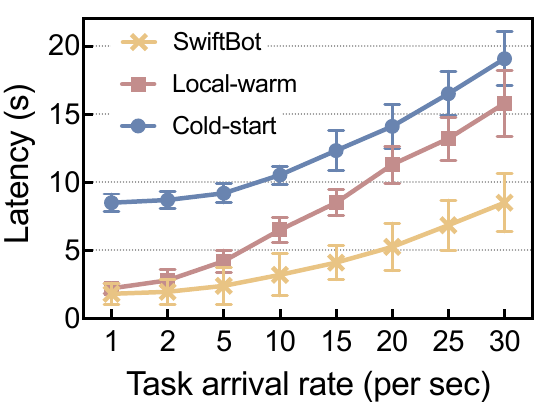}
        \caption{P99 Latency vs. Task Arrival Rate}
        \label{fig:3b}
    \end{subfigure}

    \caption{Distribution of task startup latency and the tail latency as a function of task arrival rate.}
    \label{fig:container}
\end{figure}
%%%%%%%%%%%%%%%%%%%%%%%%%%%%%%%%%%%%%%%%%%%%%%%%%%

\section{Discussion}
\label{sec:discussion}

\textbf{Scalability.} SwiftBot's Chord-based DHT overlay achieves $O(\log N)$ lookup complexity with low query latency in our testbed. However, two factors may limit larger deployments. First, DHT maintenance overhead (heartbeats, failure detection) consumes small network bandwidth with stable nodes but increases significantly under high churn. Second, lazy container state propagation (e.g., 500ms update intervals) introduces missing available containers under moderate load, rising bandwidth consumption during workload fluctuations. Future work could explore sophisticated policy for approximate availability tracking or adaptive push-pull protocols based on workload volatility.

\textbf{LLM Trade-offs.} GPT-4o's inference latency is acceptable for planning tasks executing at seconds-to-minutes timescales but prohibitive for sub-second replanning (e.g., dynamic obstacle avoidance). Fast small models (e.g., GPT-4o-mini) reduce latency but sacrifice accuracy (88.2\% vs 94.3\%), risking execution failures. API-based deployment incurs costs and raises privacy concerns. On-device quantized models (e.g., Llama) achieve quick inference but with further accuracy degradation (85.6\%), creating a three-way trade-off between latency, cost/privacy, and accuracy that requires application-specific tuning.

%\textbf{Container Migration Overhead.} Migration benefits diminish for stateful containers with large pre-loaded data. Layer-based incremental transfer could reduce data volume by 60-80\% when target nodes share base layers, while distributed shared filesystems enable metadata-only migration. Current implementation demonstrates benefits for lightweight containers; heavier workloads need further optimization.

\textbf{Resource-Constrained Platforms.} SwiftBot assumes container-capable robots, limiting applicability to embedded systems with limited RAM. Lightweight runtimes reduce overhead and WebAssembly provides millisecond startup, but severely constrained robots may require full computation offloading. The DHT architecture naturally supports offloading through migration, but the decomposer needs capability-aware planning to avoid assigning intensive subtasks to incapable platforms.
\section{Related Work}
\label{sec:related}

\textbf{Multi-Robot Collaboration.} Multi-robot collaboration entails the coordinated interaction of multiple autonomous agents to achieve objectives exceeding the capabilities of a single robot~\cite{prorok2021beyond}. Modern robots predominantly utilize the ROS 2~\cite{jo2025critical} identify that maintaining synchronization in dynamic, bandwidth-constrained environments remains a significant bottleneck for large-scale systems. Recent work explores role-based cooperation to address these challenges. Xue et al.\cite{mayya2021resilient} develop resilient task allocation that dynamically reassigns tasks in heterogeneous teams during disruptions. In communication-denied environments, specialized ``relay'' robots repair network topology~\cite{robinson2025robot, mosteo2024comm}. The field is moving toward distributed consensus approaches and AI integration for robust team performance~\cite{prorok2021beyond, zhang2025survey}.

\textbf{Human-in-the-Loop Collaboration.} The Human-in-the-Loop (HITL) paradigm involves human operators as integral parts of the control loop, allowing intervention or guidance to ensure safe and effective operation~\cite{he2025hitl}. Recent works leverage large language models (LLMs) for natural language robot control. Mandi et al.\cite{lykov2023llmbra} and LLM-MCoX~\cite{wang2025llm} translate natural language into executable behavior trees for exploration tasks. However, deploying LLMs on edge robots introduces safety risks from "hallucinations". The SAFER framework~\cite{safer2025} mitigates this via a dedicated safety agent and LLM-as-a-judge module that audit plans and enforce control-barrier constraints before execution. Furthermore, computational costs on edge devices create prohibitive latency~\cite{zhang2023large}. SwiftBot is designed to support the efficient offloading and warm-start mechanisms needed for edge deployment.

\textbf{Federated Learning and Edge Intelligence.} Federated Learning (FL) enables collaborative model training while preserving privacy by sharing gradients rather than raw sensor data. Recent FL techniques have been applied to robotic systems for distributed learning across networked agents. Gutierrez et al.~\cite{gutierrez2025fl} propose ROS 2-based FL framework for joint model training in simulation and real-world deployments. Yu et al.\cite{mofleur2024} introduce MoFLeuR to handle non-IID motion data in HRI, while Barroso et al.~\cite{barroso2025container} emphasize rapid failure recovery and live container migration. SwiftBot addresses this by replacing heavy consensus protocols with a lightweight Distributed Hash Table (DHT) overlay to achieve quick responsiveness required for robotics.

\

\section{Conclusion}
\label{sec:conclusion}

In this paper, we propose SwiftBot, a federated robotic task execution system that unifies LLM-based task decomposition with decentralized container orchestration to enable scalable and low-latency multi-robot collaboration
SwiftBot is built on three design innovations: a LLM-powered semantic task decomposition engine, a distributed warm-container layer minimizing startup latency, and a DHT-based decentralized scheduler that delivers high throughput. Evaluation on multimedia tasks validates that co-designing semantic understanding with resource management enables both flexibility and efficiency for federated robotic tasks control. In future, we will explore to support real-world large-scale heterogeneous robot tasks, optimizing lightweight LLM deployment, incorporating reinforcement learning for container pre-warming optimization, and enabling privacy-preserving edge-based LLM deployment.

\section*{Acknowledgment}

This work was supported in part by the National Science Foundation (NSF \#2436427, \#2426470, and \#2436428). The authors would like to thank reviewers for their valuable discussions and suggestions.

\bibliographystyle{IEEEtran}
%\bibliography{reference}
\bibliography{ref_new}

@article{jiang2021road,
  title={The road towards 6G: A comprehensive survey},
  author={Jiang, Wei and Han, Bin and Habibi, Mohammad Asad and Schotten, Hans D},
  journal={IEEE Open Journal of the Communications Society},
  volume={2},
  pages={334--366},
  year={2021},
  publisher={IEEE}
}

@article{prorok2021beyond,
  title={Beyond robustness: A taxonomy of approaches towards resilient multi-robot systems},
  author={Prorok, Amanda and Malencia, Matthew and Carlone, Luca and Sukhatme, Gaurav S and Sadler, Brian M and Kumar, Vijay},
  journal={arXiv preprint arXiv:2109.12343},
  year={2021}
}

@article{mayya2021resilient,
  title={Resilient task allocation in heterogeneous multi-robot systems},
  author={Mayya, Siddharth and D'Antonio, Diego S and Saldana, David and Kumar, Vijay},
  journal={IEEE Robotics and Automation Letters},
  volume={6},
  number={2},
  pages={1327--1334},
  year={2021},
  publisher={IEEE}
}

@article{zhang2023large,
  title={Large language models for human--robot interaction: A review},
  author={Zhang, Ceng and Chen, Junxin and Li, Jiatong and Peng, Yanhong and Mao, Zebing},
  journal={Biomimetic Intelligence and Robotics},
  volume={3},
  number={4},
  pages={100131},
  year={2023},
  publisher={Elsevier}
}

@article{jo2025critical,
  title={Generation of Critical Information and Sharing Mechanism for Autonomous Multi-Robot Collaboration},
  author={Jo, D. and Kwon, Y.},
  journal={IEEE Access},
  volume={13},
  pages={146856},
  year={2025},
  publisher={IEEE}
}

@article{robinson2025robot,
  title={Robot-relay: building-wide, calibration-less visual servoing with learned sensor handover networks},
  author={Robinson, Luke and De Martini, Daniele},
  journal={Autonomous Robots},
  volume={50},
  number={3},
  year={2025},
  publisher={Springer}
}

@inproceedings{mosteo2024comm,
  title={Bilayer Real Time Multi-Robot Communication Maintenance Deployment Framework for Robot Swarms},
  author={Zhang, S. and Others},
  booktitle={Journal of Physics: Conference Series},
  volume={2850},
  pages={012011},
  year={2024}
}

@article{wang2025llm,
  title={LLM-MCoX: Large Language Model-based Multi-robot Coordinated Exploration and Search},
  author={Wang, Ruiyang and Hsu, Hao-Lun and Hunt, David and Luo, Shaocheng and Kim, Jiwoo and Pajic, Miroslav},
  journal={arXiv preprint arXiv:2509.26324},
  year={2025}
}

@article{safer2025,
  title={Safety Aware Task Planning via Large Language Models in Robotics},
  author={Khan, Azal Ahmad and Andrev, Michael and Murtaza, Muhammad Ali and Aguilera, Sergio and Zhang, Rui and Ding, Jie and Hutchinson, Seth and Anwar, Ali},
  journal={arXiv preprint arXiv:2503.15707},
  year={2025}
}

@inproceedings{mofleur2024,
  title={MoFLeuR: Motion-based Federated Learning Gesture Recognition},
  author={Seyedmohammadi, S. Jamal and Sheikholeslami, S. Mohammad and Abouei, Jamshid and Mohammadi, Arash and Plataniotis, Konstantinos N.},
  booktitle={2024 IEEE 4th International Conference on Human-Machine Systems (ICHMS)},
  pages={1--6},
  year={2024},
  organization={IEEE}
}

@article{barroso2025container,
  title={A comprehensive performance evaluation of container migration strategies},
  author={Barroso, Jo{\~a}o Pedro Resende and Manacero, Aleardo and Lobato, Renata Spolon and Spolon, Roberta},
  journal={Computing},
  volume={107},
  number={2},
  year={2025},
  publisher={Springer}
}

@article{lykov2023llmbra,
  title={{LLM-BRAIn}: AI-driven Fast Generation of Robot Behaviour Tree based on Large Language Model},
  author={Lykov, Artem and Tsetserukou, Dzmitry},
  journal={arXiv preprint arXiv:2305.19352},
  year={2023}
}

@article{gutierrez2025fl,
  title={Federated Learning for Collaborative Robotics: A ROS 2-Based Approach},
  author={Gutierrez, Gerardo M and Rincon, Jaime A and Julian, Vicente},
  journal={Electronics},
  volume={14},
  number={7},
  pages={1323},
  year={2025},
  publisher={MDPI}
}

@article{zhang2025survey,
  title={A Survey on Multi-Robot Collaboration Systems: Architectures, Performances, and Applications},
  author={Zhang, Simon and Li, Zhengxiong and Yin, Yaxuan and Xu, Shuai and Chaudhary, Vipin and Xu, Hailu},
  journal={TechRxiv preprint},
  year={2025},
  note={doi:10.36227/techrxiv.176045766.60277537}
}

@article{he2025hitl,
  title={Uncertainty Comes for Free: Human-in-the-Loop Policies with Diffusion Models},
  author={He, Zhanpeng and Cao, Yifeng and Ciocarlie, Matei},
  journal={arXiv preprint arXiv:2503.01876},
  year={2025}
}

@article{devan2021wirelesshart,
  author  = {Devan, P. Arun Mozhi and Hussin, Fawnizu Azmadi and Ibrahim, Rosdiazli and Bingi, Kishore and Khanday, Farooq Ahmad},
  title   = {A Survey on the Application of WirelessHART for Industrial Process Monitoring and Control},
  journal = {Sensors},
  volume  = {21},
  number  = {15},
  pages   = {4951},
  year    = {2021},
  doi     = {10.3390/s21154951}
}

@article{kamtam2024teleop,
  author  = {Kamtam, Sidharth Bhanu and Lu, Qian and Bouali, Faouzi and Haas, Olivier C. L. and Birrell, Stewart},
  title   = {Network Latency in Teleoperation of Connected and Autonomous Vehicles: A Review of Trends, Challenges, and Mitigation Strategies},
  journal = {Sensors},
  volume  = {24},
  number  = {12},
  pages   = {3957},
  year    = {2024},
  doi     = {10.3390/s24123957}
}

@article{gielis2022communications,
  author  = {Gielis, Jennifer and Shankar, Ajay and Prorok, Amanda},
  title   = {A Critical Review of Communications in Multi-Robot Systems},
  journal = {Current Robotics Reports},
  volume  = {3},
  number  = {3},
  pages   = {213--225},
  year    = {2022},
  doi     = {10.1007/s43154-022-00090-9}
}

@article{groshev2023edgerobotics,
  author  = {Groshev, Milan and Baldoni, Gabriele and Cominardi, Luca and de la Oliva, Antonio and Gazda, Robert},
  title   = {Edge Robotics: Are We Ready? An Experimental Evaluation of Current Vision and Future Directions},
  journal = {Digital Communications and Networks},
  volume  = {9},
  number  = {1},
  pages   = {166--174},
  year    = {2023},
  doi     = {10.1016/j.dcan.2022.04.032}
}

@article{saboia2022achord,
  author  = {Saboia, Maira and et al.},
  title   = {ACHORD: Communication-Aware Multi-Robot Coordination with Intermittent Connectivity},
  journal = {IEEE Robotics and Automation Letters},
  volume  = {7},
  number  = {4},
  pages   = {10184--10191},
  year    = {2022},
  doi     = {10.1109/LRA.2022.3193240}
}

@article{choudhury2022dynamicmrta,
  author  = {Choudhury, Shushman and Gupta, Jayesh K. and Kochenderfer, Mykel J. and Sadigh, Dorsa and Bohg, Jeannette},
  title   = {Dynamic Multi-Robot Task Allocation Under Uncertainty and Temporal Constraints},
  journal = {Autonomous Robots},
  volume  = {46},
  number  = {1},
  pages   = {231--247},
  year    = {2022},
  doi     = {10.1007/s10514-021-10022-9}
}

@article{mao2024robomatrix,
  title={Robomatrix: A skill-centric hierarchical framework for scalable robot task planning and execution in open-world},
  author={Mao, Weixin and Zhong, Weiheng and Jiang, Zhou and Fang, Dong and Zhang, Zhongyue and Lan, Zihan and Li, Haosheng and Jia, Fan and Wang, Tiancai and Fan, Haoqiang and others},
  journal={arXiv preprint arXiv:2412.00171},
  year={2024}
}

@article{mtowe2025low,
  title={Low-Latency Edge-Enabled Digital Twin System for Multi-Robot Collision Avoidance and Remote Control},
  author={Mtowe, Daniel Poul and Long, Lika and Kim, Dong Min},
  journal={Sensors},
  volume={25},
  number={15},
  pages={4666},
  year={2025},
  publisher={MDPI}
}

@article{celik2024decentralized,
  title={Decentralized system synchronization among collaborative robots via 5G technology},
  author={Celik, Ali Ekber and Rodriguez, Ignacio and Ayestaran, Rafael Gonzalez and Yavuz, Sirma Cekirdek},
  journal={Sensors (Basel, Switzerland)},
  volume={24},
  number={16},
  pages={5382},
  year={2024}
}

@article{stoica2003chord,
  title={Chord: a scalable peer-to-peer lookup protocol for internet applications},
  author={Stoica, Ion and Morris, Robert and Liben-Nowell, David and Karger, David R and Kaashoek, M Frans and Dabek, Frank and Balakrishnan, Hari},
  journal={IEEE/ACM Transactions on networking},
  volume={11},
  number={1},
  pages={17--32},
  year={2003},
  publisher={IEEE}
}

@techreport{openai2024gpt4o,
  author = {OpenAI},
  title = {{GPT-4o} System Card},
  institution = {OpenAI},
  year = {2024},
  url = {https://openai.com/index/hello-gpt-4o/}
}

@techreport{anthropic2024claude35sonnet,
  author = {Anthropic},
  title = {Claude 3.5 Sonnet},
  institution = {Anthropic},
  year = {2024},
  url = {https://www.anthropic.com/news/claude-3-5-sonnet}
}

@techreport{openai2024gpt4omini,
  author = {OpenAI},
  title = {{GPT-4o mini}: Advancing Cost-Efficient Intelligence},
  institution = {OpenAI},
  year = {2024},
  url = {https://openai.com/index/gpt-4o-mini-advancing-cost-efficient-intelligence/}
}

@article{dubey2024llama,
  title = {The {Llama} 3 Herd of Models},
  author = {Dubey, Abhimanyu and others},
  journal = {arXiv preprint arXiv:2407.21783},
  year = {2024},
  url = {https://arxiv.org/abs/2407.21783}
}

@article{jiang2019task,
  title={Task planning in robotics: an empirical comparison of {PDDL}- and {ASP}-based systems},
  author={Jiang, Yuqian and Zhang, Shiqi and Khandelwal, Piyush and Stone, Peter},
  journal={Frontiers of Information Technology \& Electronic Engineering},
  volume={20},
  number={3},
  pages={363--373},
  year={2019},
  publisher={Springer}
}

@article{mon2025embodied,
  title={Embodied large language models enable robots to complete complex tasks in unpredictable environments},
  author={Mon-Williams, Ruaridh and Li, Gen and Long, Ran and Du, Wenqian and Lucas, Christopher G},
  journal={Nature Machine Intelligence},
  pages={1--10},
  year={2025},
  publisher={Nature Publishing Group UK London}
}

@article{you2025construction,
  title={Construction Robotics in Extreme Environments: From Earth to Space},
  author={You, Ke and Zhou, Cheng and Ding, Lieyun and Wang, Yuxiang},
  journal={Engineering},
  year={2025},
  publisher={Elsevier}
}

@article{matos2025efficient,
  title={Efficient multi-robot path planning in real environments: a centralized coordination system},
  author={Matos, Diogo Miguel and Costa, Pedro and Sobreira, H{\'e}ber and Valente, Antonio and Lima, Jos{\'e}},
  journal={International Journal of Intelligent Robotics and Applications},
  volume={9},
  number={1},
  pages={217--244},
  year={2025},
  publisher={Springer}
}

@article{bhaskaran2025comprehensive,
  title={A Comprehensive Study of Resource Provisioning and Optimization in Edge Computing.},
  author={Bhaskaran, Sreebha and Muthuraman, Supriya},
  journal={Computers, Materials \& Continua},
  volume={83},
  number={3},
  year={2025}
}

@online{harrison2020avoiding,
  author = {Harrison, Josh},
  title = {Avoiding cold starts on {AWS Lambda} for a long-running {API} request},
  year = {2020},
  month = {October},
  day = {22},
  url = {https://towardsdatascience.com/avoiding-cold-starts-on-aws-lambda-for-a-long-running-api-request-15b8194f2e01/},
}

@article{golec2024cold,
  title={Cold start latency in serverless computing: A systematic review, taxonomy, and future directions},
  author={Golec, Muhammed and Walia, Guneet Kaur and Kumar, Mohit and Cuadrado, Felix and Gill, Sukhpal Singh and Uhlig, Steve},
  journal={ACM Computing Surveys},
  volume={57},
  number={3},
  pages={1--36},
  year={2024},
  publisher={ACM New York, NY}
}

@inproceedings{yang2024plug,
  title={Plug in the safety chip: Enforcing constraints for llm-driven robot agents},
  author={Yang, Ziyi and Raman, Shreyas S and Shah, Ankit and Tellex, Stefanie},
  booktitle={2024 IEEE International Conference on Robotics and Automation (ICRA)},
  pages={14435--14442},
  year={2024},
  organization={IEEE}
}

@inproceedings{kannan2024smart,
  title={Smart-llm: Smart multi-agent robot task planning using large language models},
  author={Kannan, Shyam Sundar and Venkatesh, Vishnunandan LN and Min, Byung-Cheol},
  booktitle={2024 IEEE/RSJ International Conference on Intelligent Robots and Systems (IROS)},
  pages={12140--12147},
  year={2024},
  organization={IEEE}
}

@inproceedings{karli2024alchemist,
  title={Alchemist: Llm-aided end-user development of robot applications},
  author={Karli, Ulas Berk and Chen, Juo-Tung and Antony, Victor Nikhil and Huang, Chien-Ming},
  booktitle={Proceedings of the 2024 ACM/IEEE International Conference on Human-Robot Interaction},
  pages={361--370},
  year={2024}
}

@article{chen2023scalable,
  title={Scalable Multi-Robot Collaboration with Large Language Models: Centralized or Decentralized Systems?},
  author={Chen, Yongchao and Arkin, Jacob and Zhang, Yang and Roy, Nicholas and Fan, Chuchu},
  journal={arXiv preprint arXiv:2309.15943},
  year={2023}
}

@inproceedings{oakes2018sock,
  title={{SOCK}: Rapid Task Provisioning with Serverless-Optimized Containers},
  author={Oakes, Edward and Yang, Leon and Zhou, Dennis and Houck, Kevin and Harter, Tyler and Arpaci-Dusseau, Andrea and Arpaci-Dusseau, Remzi},
  booktitle={2018 USENIX Annual Technical Conference (USENIX ATC 18)},
  pages={57--70},
  year={2018}
}

@article{nasir2024cold,
  title={Cold Start Latency in Serverless Computing: A Systematic Review, Taxonomy, and Future Directions},
  author={Nasir, Wajid and Azim, Akramul and others},
  journal={ACM Computing Surveys},
  year={2024},
  publisher={ACM}
}

@article{gupta2025performance,
  title={Performance Characterization of Containers in Edge Computing},
  author={Gupta, Ragini and Nahrstedt, Klara},
  journal={arXiv preprint arXiv:2505.02082},
  year={2025}
}

@techreport{soomro2012ucf101,
  author       = {Soomro, Khurram and Zamir, Amir Roshan and Shah, Mubarak},
  title        = {UCF101: A Dataset of 101 Human Action Classes From Videos in The Wild},
  institution  = {Center for Research in Computer Vision (UCF)},
  year         = {2012},
  note         = {Technical Report CRCV-TR-12-01},
  url          = {https://www.crcv.ucf.edu/data/UCF101.php}
}

@inproceedings{panayotov2015librispeech,
  author    = {Panayotov, Vassil and Chen, Guoguo and Povey, Daniel and Khudanpur, Sanjeev},
  title     = {LibriSpeech: An ASR Corpus Based on Public Domain Audio Books},
  booktitle = {2015 IEEE International Conference on Acoustics, Speech and Signal Processing (ICASSP)},
  year      = {2015},
  pages     = {5206--5210},
  doi       = {10.1109/ICASSP.2015.7178964}
}

@article{wen2023survey,
  title={A survey on federated learning: challenges and applications},
  author={Wen, Jie and Zhang, Zhixia and Lan, Yang and Cui, Zhihua and Cai, Jianghui and Zhang, Wensheng},
  journal={International journal of machine learning and cybernetics},
  volume={14},
  number={2},
  pages={513--535},
  year={2023},
  publisher={Springer}
}

@inproceedings{guo2025lightllm,
  title={LightLLM-Enhanced Multi-Robot Collaboration with Human-in-the-Loop},
  author={Guo, Jiaming and Qi, Zhichun and Zhao, Xiaokuan and Xu, Shuai and Xu, Hailu},
  booktitle={2025 8th International Conference on Robotics, Control and Automation Engineering (RCAE)},
  pages={27--32},
  year={2025},
  organization={IEEE}
}

@inproceedings{zhang2025towards,
  title={Towards a lightweight platform for human-robot interaction in federated edge and iot environments},
  author={Zhang, Simon and Li, Zhengxiong and Qin, Xin and Xu, Hailu},
  booktitle={Proceedings of the 3rd International Workshop on Human-Centered Sensing, Modeling, and Intelligent Systems},
  pages={110--113},
  year={2025}
}

\end{document}